\documentclass{article}
\usepackage{spconf,amsmath,graphicx}
\usepackage{amssymb}

\title{Identification of Dynamic functional brain network states Through Tensor Decomposition}
\name{Arash Golibagh Mahyari, Selin Aviyente \thanks{This work was in part supported by the National Science Foundation
under Grant No. CCF-1218377.}}
\address{mahyari@msu.edu, aviyente@egr.msu.edu \\
Department of Electrical and Computer Engineering \\
Michigan State University \\
East Lansing, MI 48823 USA}

\begin{document}

%
\maketitle
\begin{abstract}
With the advances in high resolution neuroimaging, there has been a growing interest in the detection of functional brain connectivity. Complex network theory has been proposed as an attractive mathematical representation of functional brain networks. However, most of the current studies of functional brain networks have focused on the computation of graph theoretic indices for static networks, i.e. long-time averages of connectivity networks. It is well-known that  functional connectivity is a dynamic process and the construction and reorganization of the networks is key to understanding human cognition. Therefore, there is a growing need to track dynamic functional brain networks and identify time intervals over which the network is quasi-stationary. In this paper, we present a tensor decomposition based method to identify temporally invariant 'network states' and find a common topographic representation for each state. The proposed methods are applied to electroencephalogram (EEG) data during the study of error-related negativity (ERN).
\end{abstract}
\section{Introduction}
\label{sec:intro}

With the advance in noninvasive imaging modalities such as fMRI, EEG, and MEG, it is important to develop computational methods capable of giving a succinct description of the functional brain networks \cite{he2013grand}. Functional connectivity describes the coordinated activation of segregated groups of neurons. Traditionally, functional connectivity has been quantified through linear  \,\cite{rubinov2010complex} and nonlinear measures \,\cite{pereda2005nonlinear}. Synchronization of neuronal oscillations has been suggested as one plausible mechanism in the interaction of spatially distributed neural populations and has been quantified using phase synchrony \cite{varela2001brainweb}. Although phase synchrony is successful at quantifying pairwise interactions \cite{bullmore2009complex}, it cannot completely describe the complex relationship between function and organization of the brain.  Recently, research in the area of complex networks has led to fundamental insights into the organization of the healthy and diseased brain \cite{bullmore2009complex,stam2007graph}. However, the current studies are limited to the analysis of static brain networks obtained through averaging long-term functional connectivity and thus, neglect possible time-varying properties of the topologies.
\vspace{-1mm}

There is growing evidence that functional networks dynamically reorganize and coordinate on millisecond scale for the execution of mental processes \cite{dimitriadis2013quantization}. For this reason, there has been an interest in characterizing the dynamics of functional networks using high temporal resolution EEG recordings. The early work in this area was an extension of static network analysis to the dynamic case by extracting graph theoretic features from graphs across time and tracking the evolution of these parameters \cite{dimitriadis2010tracking}. However, these approaches lose the spatial information provided by the graphs and cannot identify which parts of the brain contributed to the observed changes in the network. More recently, a "network state" framework has been proposed \cite{betzel2012synchronization,dimitriadis2013quantization}, where each state is defined as periods of time during which the network topology is quasi-stationary. In this paper, we adopt this framework for tracking the topology of brain networks across time and representing each state with a common topographic map. The current work differs from the existing approaches in a couple of ways. First, we take into account the full structure of the network at each time point. Second, we consider extracting network states common across time and subjects unlike current work which considers individual subjects. Finally, the current work offers a compressed spatial representation of each network state through tensor-tensor projection unlike current approaches which use averaging. Tensor-to-tensor projection proposed in this paper projects the information across subjects and time into a lower dimensional 'signal' subspace whereas averaging assigns equal weights to all subjects.

It is also important to note that the proposed framework is closely tied to dynamic network tracking. The most common approaches to network tracking have been to identify anomalies using subspace projection methods such as in \cite{lakhina2004diagnosing,miller2011eigen} or through sliding window estimates with time independent or dependent weighting factors \cite{tong2008,hero2011shrinkage}. More recently, adaptive evolutionary clustering \cite{xu2011adaptive} was proposed to track the cluster changes over time. However, all of these methods either detect time points where events of interest happen, or find different clustering structures at each time point. In this paper, we propose a comprehensive framework that first identifies time intervals during which the network topology is stationary and then summarizes each interval by a single lower dimensional network through tensor-tensor projection.

\section{Background}
\label{sec:format}

\subsection{Time-Varying Network Construction}
\label{ssec:subhead}

The time-varying functional brain networks are constructed from multichannel EEG data with the nodes corresponding to the different brain regions and the edges to the connectivity between these regions. In this paper, we quantify the connectivity using a recently introduced phase synchrony measure based on RID-Rihaczek distribution \cite{aviyente2011time}. The first step in quantifying phase synchrony is to estimate the time and frequency dependent phase, $\Phi_i(t,\omega)$, of a signal, $\mathrm{s_i}$, $\mathrm{arg}\left[\frac{C_i(t,\omega)}{|C_i(t,\omega)|}\right]$,
where $C_i(t,\omega)$ is the complex RID-Rihaczek distribution \footnote{The details of the RID-Rihaczek distribution and the corresponding synchrony measure are given in \cite{aviyente2011time}.}:
\begin{equation}
\footnotesize
C_i(t,\omega)=\int \! \int \! \!\underbrace{\exp\left(-\frac{(\theta
\tau)^{2}}{\sigma}\right)}_{\text{\tiny{Choi-Williams} \hspace{0.5mm}kernel}}
\underbrace{\exp(j\frac{\theta \tau}{2})}_{\text{\tiny{Rihaczek}
\hspace{0.5mm}kernel}}\!A_i(\theta,\tau) e^{-j(\theta t+\tau
\omega)} d\tau d\theta
\end{equation}\newline
and $A_i(\theta,\tau)=\int s_i(u+\frac{\tau}{2})s_i^{*}(u-\frac{\tau}{2})e^{j\theta u} du$ is the ambiguity function of the signal ${\it{s}}_i$. The phase synchrony between nodes $i$ and $j$ at time $t$ and frequency $\omega$ is computed using Phase Locking Value (PLV):
\begin{equation}
PLV_{ij}(t,\omega) = \frac{1}{L}\left|\sum_{k=1}^L \exp{\left(j\Phi_{ij}^k(t,\omega)\right)}\right|
\label{eq:plv}
\end{equation}
where $L$ is the number of trials and $\Phi_{ij}^{k}(t,\omega)=|\Phi^{k}_i(t,\omega)-\Phi^{k}_j(t,\omega)|$  is the phase difference between the two channels for the $k^{th}$ trial.

Once the pairwise synchrony values are computed at each time and frequency point, we can construct a time-varying graph $\{{\bf{G}}(t)\}_{t=1,2,\dots,T}$ with
\begin{equation}
G_{ij}(t) = \frac{1}{\Omega}\sum_{\omega=\omega_a}^{\omega_b}PLV_{ij}(t,\omega)
\label{eq:Graph}
\end{equation}
\noindent where $G_{ij}(t)\in [0,1]$ represents the connectivity strength between the nodes $i$ and $j$ within the frequency band of interest, $[\omega_a,\omega_b]$, and $\Omega$ is the number of frequency bins in that band.

Therefore, $\{{\bf{G}}(t)\}_{t=1,2,\dots,T}$ is a time series of $N\times N$ weighted and undirected graphs corresponding to the functional connectivity network at time $t$ for a fixed frequency band, where $T$ is the total number of time points and $N$ is the number of nodes within the network.

\subsection{Tensor Subspace Analysis}
\label{ssec:subhead}

Linear data models such as Principal Component Analysis (PCA) and Independent Component Analysis (ICA) are widely used for the decomposition of matrices. Depending on the criteria, different types of basis vectors are extracted and appropriate lower dimensional features are determined through projection. Multiway data analysis extends these linear methods to capture multilinear structures and underlying correlations in higher-order datasets, also known as tensors. Some exemplary methods include PARAFAC, Tucker decomposition, and Higher-Order Singular Value Decomposition (HOSVD) \cite{acar2009unsupervised,de2000multilinear}.

The Tucker decomposition is a higher order generalization of Singular Value Decomposition (SVD) \cite{acar2009unsupervised}. Let ${\mathcal{X}} \in {\mathbb{R}}^{m_1 \times m_2 ... \times m_d}$ be a ${\it{d}}$-mode array, then its Tucker decomposition can be expressed as: 
\begin{equation}
\begin{array}{c}
{\mathcal{X}} = {\mathcal{C}} \times_1 {\bf{U}}^{(1)} \times_2 {\bf{U}}^{(2)} ... \times_d {\bf{U}}^{(d)} + {\mathcal{E}} \\
=\sum_{i_1,i_2,...,i_d}{{\mathcal{C}}_{i_1,i_2,...,i_d} \left ({\bf u}^{(1)}_{i_1} \circ {\bf u}^{(2)}_{i_2} ... \circ {\bf u}^{(d)}_{i_d}   \right ) + {\mathcal{E}}_{i_1,i_2,...,i_d}}
\end{array}
\label{eq:parafac}
\end{equation}
\noindent where ${\mathcal{C}} \in {\mathbb{R}}^{r_1 \times r_2 ... \times r_d}$ is the core tensor, and ${\bf{U}}^{(1)} \in {\mathbb{R}}^{m_1 \times r_1}$, ${\bf{U}}^{(2)} \in {\mathbb{R}}^{m_2 \times r_2}$, ..., ${\bf{U}}^{(d)} \in {\mathbb{R}}^{m_d \times r_d}$, where $r_1 \leq m_1, r_2 \leq m_2, ...,r_d \leq m_d$, are the projection matrices whose columns are orthogonal. ${\mathcal{E}} \in {\mathbb{R}}^{m_1 \times m_2 ... \times m_d}$  is the residual error, and $\times_k$ is the product of a tensor and a matrix along mode-{\it{k}}. Reconstruction of the original tensor, $\tilde {\mathcal{X}} \in {\mathbb{R}}^{m_1 \times m_2 ... \times m_d}$, from a lower dimensional representation is obtained as: 

\begin{equation}
\begin{array}{c}
\tilde {\mathcal{X}} = {\mathcal{X}} \times_1 {\left ( {\bf{U}}^{(1)}{\bf{U}}^{(1)^{\dag}} \right )} \times_2 {\left ( {\bf{U}}^{(2)}{\bf{U}}^{(2)^{\dag}} \right )} ... \times_d {\left ( {\bf{U}}^{(d)}{\bf{U}}^{(d)^{\dag}} \right )}
\end{array}
\label{eq:ten_app}
\end{equation}
\noindent where ${\dag}$ is the transpose of the matrix.

\section{The Proposed Method}
\label{sec:pagestyle}

\subsection{Temporal Tracking for Network State Identification}

In the proposed work, the time-varying functional connectivity graphs across subjects are considered as a 4-mode tensor ${\mathcal{G}} \in {\mathbb{R}}^{N \times N \times T \times S}$ constructed as channel $\times$ channel $\times$ time $\times$ subject, with $N$ being the number of channels, $T$ the total number of time points and $S$ the number of subjects. The Tucker decomposition of this connectivity tensor yields:

\begin{equation}
\begin{array}{c}
{\mathcal{G}} = {\mathcal{C}} \times_1 {\bf{U}}^{(1)} \times_2 {\bf{U}}^{(2)} \times_3 {\bf{U}}^{(3)} \times_4 {\bf{U}}^{(4)} + {\mathcal{E}}
\label{eq:graph_tensor}
\end{array}
\end{equation}
\noindent where ${\mathcal{C}} \in {\mathbb{R}}^{{ N} \times { N} \times T \times { S}}$ is the core tensor, and ${\mathcal{E}}^{N \times N \times T \times S}$ is the residual error.

To obtain an approximation of ${\mathcal{G}}$, $\tilde{{\mathcal{G}}} \in {\mathbb{R}}^{{N} \times {N} \times T \times {S}}$, we first consider the full Tucker decomposition with ${\bf{U}}^{(1)} \in {\mathbb{R}}^{N \times N}$, ${\bf{U}}^{(2)} \in {\mathbb{R}}^{N \times N}$, ${\bf{U}}^{(3)} \in {\mathbb{R}}^{T \times T}$, and ${\bf{U}}^{(4)} \in {\mathbb{R}}^{S \times S}$.
The singular values along each mode are ordered by fixing the index of all of the other modes to $1$. Since first singular values along each mode represent the largest variance of the data along that mode, we choose that to order the remaining mode. To get the approximation tensor, the appropriate number of singular vectors along first and second modes ${\bar N}$ is defined as ${\bar N}=j_k$, where $j_k$ is the highest index for which $\left | {\mathcal{C}}_{j_k,1,1,1} \right | \geq 0$. Similarly, the number of singular vectors along the fourth mode is ${\bar S}=s_k$, where $s_k$ is the highest index for which $\left | {\mathcal{C}}_{1,1,1,s_k} \right | \geq 0$. The time mode is not projected to a lower dimensional space since all time points are necessary to identify the exact boundaries of the network states. The lower dimensional projection matrices are defined as: ${\tilde{\bf{U}}^{(1)}} = [{\bf{u}}_1^{(1)} {\bf{u}}_2^{(1)} ... {\bf{u}}_{\tilde{N}}^{(1)}]$, ${\tilde{\bf{U}}^{(2)}} = [{\bf{u}}_1^{(2)} {\bf{u}}_2^{(2)} ... {\bf{u}}_{\tilde{N}}^{(2)}]$, ${\tilde{\bf{U}}^{(4)}} = [{\bf{u}}_1^{(4)} {\bf{u}}_2^{(4)} ... {\bf{u}}_{\tilde{S}}^{(4)}]$.

The reconstructed tensor $\tilde{{\mathcal{G}}} \in {\mathbb{R}}^{N \times N \times T \times S}$ is obtained as:
\begin{equation}
\begin{array}{c}
\tilde {\mathcal{G}} = {\mathcal{G}} \times_1 {\left ( \tilde{\bf{U}}^{(1)}\tilde{\bf{U}}^{(1)^{\dag}} \right )} \times_2 {\left ( \tilde{\bf{U}}^{(2)}\tilde{\bf{U}}^{(2)^{\dag}} \right )} \\
\hspace{15mm} \times_3 {\left ( \tilde{\bf{U}}^{(3)}\tilde{\bf{U}}^{(3)^{\dag}} \right )} \times_4 {\left ( \tilde{\bf{U}}^{(4)}\tilde{\bf{U}}^{(4)^{\dag}} \right )}.
\label{eq:approximation_graph}
\end{array}
\end{equation}
The 4-mode approximation tensor $\tilde {\mathcal{G}}$ can be written as a sequence of 3-mode tensors $\tilde {\mathcal{G}}_t ; t=1,2,...,T$. To detect the boundaries of network states, we propose a new temporal clustering algorithm. Unlike regular data clustering, the proposed method considers both the similarity of the lower dimensional representation of the networks as well as their closeness in time. The similarity of two networks at time $t_{1}$ and $t_{2}$ is quantified through a cosine similarity metric between $\tilde {\mathcal{G}}_{t_1}$ and $\tilde {\mathcal{G}}_{t_2}$ as follows:
\begin{equation}
\begin{array}{c}
\Delta (t_1,t_2)= {{\left < \tilde{{\mathcal{G}}}_{t_1} , \tilde{{\mathcal{G}}}_{t_2} \right > } \over {{ \parallel \tilde{{\mathcal{G}}}_{t_1} \parallel}{ \parallel \tilde{{\mathcal{G}}}_{t_2} \parallel}}} ; t_1,t_2=1,2,...,T
\label{eq:similarity_eq}
\end{array}
\end{equation}
\noindent where ${\left < a , b \right > }$ is the inner product of $a$ and $b$, and ${ \parallel a \parallel}$ is the Frobenius Norm.
Similarly, the temporal closeness between two graphs is quantified as $\Theta(t_1,t_2) = e^{-{( t_1-t_2 )^2} \over { 2\sigma^2}}; t_1,t_2=1,2,...,T$, where $\sigma$ is a parameter which determines the weighting for different time separations, and depends on the sampling frequency.

The combined similarity matrix is defined as:
\begin{equation}
\Psi(t_1,t_2) = \lambda \Theta(t_1,t_2)+(1-\lambda)\Delta(t_1,t_2) ; t_1,t_2=1,2,...,T
\end{equation}
\noindent where $\lambda \in (0,1)$ determines the trade-off between tensor similarity and time proximity. This similarity matrix is input to a standard spectral clustering algorithm combined with {\it{k}}-means to identify the boundaries of the network states \cite{von2007tutorial}.
\subsection{Topographic Compression for Network State Representation}
Once the time boundaries of the different network states are identified, each state has to be summarized with a single topographic map. Previously, this was commonly addressed by averaging the edges over the time interval \cite{mahyari2013GSIP}. This method has the drawback of emphasizing all of the edges equally and resulting in very dense network representations.

For a given time interval $(T_1,T_2)$ and the 3-mode tensor sequence corresponding to this interval ${\mathcal{G}}_{T_1},{\mathcal{G}}_{T_{1}+1}, {\mathcal{G}}_{T_{1}+2},...,{\mathcal{G}}_{T_{2}}$, the goal is to extract the topographic map $\hat{\mathcal{G}} \in {\mathbb{R}}^{N \times N}$ which best represents that network state. The 3-mode tensors corresponding to the time interval $(T_1,T_2)$ can be rewritten as a 4-mode tensor by taking the time modality into account ${\mathcal{G^\prime}} \in \mathbb{R}^{N \times N \times (T_2-T_1+1) \times S}$, and decomposed using the full Tucker decomposition, ${\bf{U}}^{{\prime}^{(1)}} \in \mathbb{R}^{N \times N}, {\bf{U}}^{{\prime}^{(2)}} \in \mathbb{R}^{N \times N}, {\bf{U}}^{{\prime}^{(3)}} \in \mathbb{R}^{(T_2-T_1+1) \times (T_2-T_1+1)}, {\bf{U}}^{{\prime}^{(4)}} \in \mathbb{R}^{S \times S} $, similar to Equation~\ref{eq:graph_tensor}.

In order to summarize the subject information to find a general unique model which fits all subjects, the 4-mode tensor ${\mathcal{G^\prime}} \in \mathbb{R}^{N \times N \times (T_2-T_1+1) \times S}$ is projected by the singular vector ${\bf{u}}^{{\prime}^{(4)}}_l$ corresponding to the $l^{th}$ largest singular value in this mode. Likewise, to summarize the time information of the resulting 3-mode tensor, it is projected to the singular vector ${\bf{u}}^{{\prime}^{(3)}}_k$ corresponding to the $k^{th}$ largest singular value of the time mode, $\hat{\mathcal{G}}= {{\mathcal{G^\prime}}} \times_3 {\bf{u}}^{{\prime}^{(3)}}_k \times_4 {\bf{u}}^{{\prime}^{(4)}}_l$. The values of k and l are usually equal to 1 but may change depending on the data.

\section{Experimental Results}
\label{sec:typestyle}

\subsection{EEG Data}
The proposed framework is applied to a set of EEG data containing the error-related negativity (ERN) \footnote{We thank Dr. Edward Bernat from the University of Maryland for sharing his EEG dataset with us.}. The ERN is a brain potential response that occurs following performance errors in a speeded reaction time task usually 25-75 ms after the response \cite{Hall}. Previous work \cite{cavanagh2009prelude} indicates that there is increased coordination between the lateral prefrontal cortex (lPFC) and medial prefrontal cortex (mPFC) within the theta frequency band (4-8 Hz) and ERN time window (25-75 ms), supporting the idea that frontal and central electrodes are functionally integrated during error processing. EEG data from $62$-channels was collected in accordance with the $10$/$20$ system on a Neuroscan Synamps2 system (Neuroscan, Inc.). A speeded-response flanker task was employed, and response-locked averages were computed for each subject. All EEG epochs were converted to current source density (CSD) using published methods \cite{kayser2006principal}. Data were averaged across trials ($\sim 200$ trials) for the purpose of ERN and time-frequency analysis.In this paper, we analyzed data from 91 subjects corresponding to the error responses.

\subsection{Network State Identification and Summarization}

\begin{figure}[htb]
  \centering
  \centerline{\includegraphics[scale=0.44]{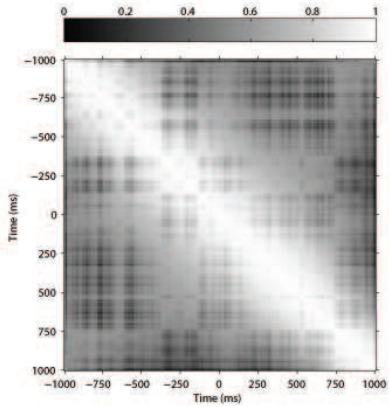}}
\caption{Similarity matrix, $\Psi$, computed for 2 seconds of EEG data across 91 subjects in the theta frequency band.}
\label{fig:psi_matrix}
\end{figure}

The connectivity matrices are constructed by computing the pairwise average  PLV between 62 channels in the theta frequency band for all time (2 seconds) and all subjects using Eq.~\ref{eq:plv}. The time-varying graphs $\{{\bf{G}}(t)\}_{t=1,2,\dots,T}$ for all subjects and all time will be treated as a 4-mode tensor, which is decomposed using Tucker decompostion. The approximation tensors $\tilde {\mathcal{G}}_t ; t=1,2,...,T$ with ${\bar N}=2, {\bar S}=3$ are used to obtain the $256 \times 256$ similarity matrix $\Psi$. The matrix $\Psi$ is computed with $\lambda=0.4$ and $\sigma=2500$ as shown in Fig.~\ref{fig:psi_matrix}. The values of $\lambda$ and $\sigma$ are empirically chosen to obtain the best separation between clusters.
Once the matrix $\bf{\Psi}$ is obtained, the critical time points are detected using the spectral clustering with $K=5$. The number of clusters $K$ is selected based on the eigenspectrum of the similarity matrix \cite{ng2002spectral}. The detected time intervals are $ (-1000,-703)$ {\it{ms}}, $(-703,-132)$ {\it{ms}}, $(-132,188)$ {\it{ms}},$(188,736)$ {\it{ms}},$(736,1000)$ {\it{ms}}.

%

As expected, the first two time intervals correspond to the prestimulus part where there is less change in the network configuration. The third time interval $(-132ms, 188ms)$ is of particular interest since it includes the time interval right before a response is made as well as the ERN interval $(0-150ms)$. This is extracted as a separate network state indicating a reorganization of the functional network configuration. Similarly, the time interval $(188,736)ms$ contains the P300 event which is expected to result in a distinct topographic map. In this paper, we will focus on extracting the common topographic map for the time interval $(-132,188)$ $ms$ since it coincides with the ERN \cite{cavanagh2009prelude}. To obtain a single network representation for this time interval, we selected $k=l=2$ instead of the singular vectors corresponding to the highest singular values since the projection to the subspace spanned by the largest singular value mostly contains edges between physically adjacent nodes. This is a side effect of volume conduction affecting PLV values and does not convey the actual long-range relationships we are interested in.

\begin{figure}[htb]
  \centering
  \centerline{\includegraphics[width=7cm] {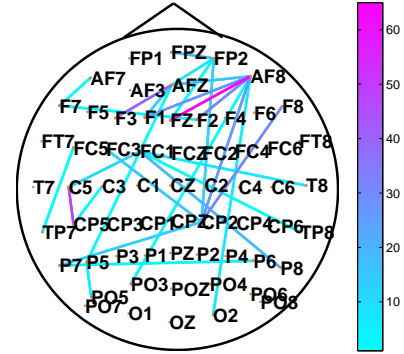}}
\caption{The topographic map with the most signifcant edges of the compressed connectivity matrix for the network state $(-132ms,188ms)$.}\label{fig:significant_edges}
\end{figure}

In order to show the most significant edges in the summarized graph, the edges with values in the top $1\%$ are selected and plotted in Fig.~\ref{fig:significant_edges}. Most of the significant edges are in the frontal and central areas where node $AF8$ acts as a hub, with the highest degree equal to $59$. $C1$ has the second highest degree (53) followed by $FP2$ (52). These nodes correspond to the right lateral prefrontal cortex (lPFC) and medial prefrontal cortex (mPFC) in accordance with previous findings which indicate increased synchronization between these regions during ERN \cite{cavanagh2009prelude}.

\section{Conclusions}
\label{sec:illust}

In this paper, we proposed a tensor-based method for monitoring dynamic functional connectivity networks to identify quasi-stationary network states and represent the common topographic distribution of each state. Network states were identified computing a similarity metric which takes both the similarity of the reconstructed tensors and their proximity in time to obtain a partitioning of the dynamic networks into contiguous time intervals. After identifying the boundaries of the network states, a topographical map for each time interval was obtained by a tensor-tensor projection. The application of the proposed algorithm to ERN data yields time intervals that closely correspond to events of interest and topographic maps that are consistent with previous hypotheses with regard to error monitoring in the brain. The proposed method is time consuming for large data sets due to the complexity of Tucker decomposition.

Future work will consider extensions of this framework to partitioning time and frequency dependent connectivity networks by considering higher order tensor representations. Moreover, the choice of optimal parameters ($\lambda$ and $\sigma$) will be considered using cost functions such as modularity for evaluating the quality of the different partitions.


\vfill\pagebreak



\bibliographystyle{IEEEbib}
\bibliography{strings_EEG,refs_EEG}

\end{document}